\documentclass[11pt]{article}

\usepackage[a4paper,margin=1in]{geometry}
\usepackage{amsmath,amssymb,amsthm}
\usepackage{booktabs}
\usepackage{graphicx}
\usepackage{hyperref}
\usepackage{url}
\usepackage{framed}
\usepackage{enumitem}
\usepackage{authblk}
\usepackage{xcolor}
\usepackage{tikz}
\usetikzlibrary{arrows.meta, positioning, fit, backgrounds, calc, shapes.geometric, shapes.symbols, shapes.misc}

\hypersetup{
  colorlinks=true,
  linkcolor=black,
  urlcolor=blue,
  citecolor=black
}

\newtheorem{proposition}{Proposition}

\newcommand{\LHB}{\textsc{LongHorizon-Bench}}

\title{Stateless Decision Memory for Enterprise AI Agents}

\author{Vasundra Srinivasan}
\affil{AI Architect, Author---\textit{Data Engineering for Multimodal AI} (O'Reilly), Stanford School of Engineering}
\date{April 2026}

\begin{document}
\maketitle

\begin{abstract}
Enterprise deployment of long-horizon decision agents in regulated domains (underwriting, claims adjudication, clinical review, tax examination) is dominated by retrieval-augmented pipelines despite the research literature producing a decade of increasingly sophisticated stateful memory architectures. We argue this reflects a hidden requirement the research evaluation has not surfaced: regulated deployment is load-bearing on four systems properties (deterministic replay, auditable rationale, multi-tenant isolation, and statelessness for horizontal scale). Stateful memory architectures violate these properties by construction, and the margin of violation compounds as deployment matures. We propose \textbf{Deterministic Projection Memory} (DPM), an architecture that treats agent memory as an append-only event log plus a single task-conditioned projection at decision time. On ten regulated decisioning cases at three memory budgets, DPM matches summarization-based memory at generous budgets (no statistically significant difference on any of four decision-alignment axes at moderate or loose budgets; $n{=}10$, paired permutation) and substantially outperforms it when memory budget binds: at a 20$\times$ compression ratio, DPM improves factual precision by $+0.52$ (Cohen's $h{=}1.17$, $p{=}0.0014$) and reasoning coherence by $+0.53$ ($h{=}1.13$, $p{=}0.0034$). DPM is additionally $7$--$15\times$ faster than the stateful baseline because it makes one LLM call at decision time instead of $N$ calls across the trajectory. A determinism study of 10 replays per case at temperature zero confirms that both conditions inherit residual API-level nondeterminism, but that the architectural asymmetry is systemic: DPM exposes one nondeterministic projection call, summarization-based memory exposes $N$ compounding calls across the trajectory. We conclude with TAMS, a practitioner heuristic for architecture selection, and a failure analysis of stateful memory under enterprise operating conditions. The contribution is not a new decision-accuracy win. It is the argument that statelessness is the load-bearing property explaining enterprise's preference for weaker but replayable retrieval pipelines, and that DPM demonstrates this property is attainable without the decisioning penalty retrieval pays.
\end{abstract}

\section{Introduction}
\label{sec:intro}

A visible gap separates the agent-memory research literature from enterprise deployment of long-horizon decision agents. Academic memory architectures have proliferated: typed routing, schema-anchored storage, hierarchical consolidation, graph-based agent memory, belief-state tracking, multi-agent memory systems \cite{memgpt2024, hymem2026, hmem2025, gam2026, mirix2025, timem2026, amem2025}. Enterprise AI deployment instead runs overwhelmingly on retrieval-augmented generation (RAG) with a vector database and chunked context assembly. This is conspicuous because RAG is straightforwardly weaker than the best memory architectures on the evaluations those architectures report.

We propose that the gap is explained by a set of systems properties the research evaluation has underweighted. A regulated decisioning system (a mortgage-underwriting agent, a claims adjudication agent, a clinical prior-authorization agent) runs in an operating environment whose requirements do not reduce to decision accuracy. The agent's memory substrate must support deterministic replay so that a denied applicant can be re-scored and the same decision justified; it must expose an auditable rationale trail a regulator, internal audit function, or court can inspect; it must provide multi-tenant isolation so that one applicant's documents cannot leak into the decision on another applicant; and it must be stateless in the sense of horizontal scalability, because an enterprise system serving thousands of concurrent decisions cannot bottleneck on a shared mutable memory. Retrieval pipelines happen to support these properties as side-effects of their architectural simplicity. Sophisticated stateful memory architectures violate at least one of them by construction.

The paper's contribution is to treat this observation as a systems claim and to evaluate it. We define Deterministic Projection Memory (DPM), a deliberately simple architecture designed to inherit retrieval's operating properties while matching summarization-based memory's decision quality. DPM treats the trajectory as an append-only immutable event log. At decision time it performs a single task-conditioned projection at temperature zero: a structured extraction of facts, reasoning points, and compliance notes within a budget. The log is the single source of truth. The projection is pure, in the functional sense; a replay from the same log under the same model version produces (up to residual API nondeterminism) the same memory view.

The empirical question is whether statelessness comes at a decisioning cost. We run DPM against summarization-based memory on ten regulated decisioning cases at three memory budgets, scored on the four decision-alignment axes from \cite{alignmentpaper}. We find: at generous budgets there is no significant difference on any axis; at tight budgets DPM is strictly better on factual precision and reasoning coherence with large effect sizes. A determinism study measures the byte-level stability of the projected memory across ten replays per case, showing that both conditions inherit residual API stochasticity at temperature zero, but that DPM exposes a single nondeterministic call while summarization exposes $N$ calls, one per consolidation step. We present TAMS, a simple task-property decision rule for architecture selection, and a failure analysis of stateful memory under enterprise operating conditions.

Contributions. (i) A named systems architecture (DPM) designed against the four enterprise properties rather than against decision-accuracy benchmarks. (ii) An empirical comparison against the strongest stateful baseline across a $3{\times}4{\times}2$ design showing DPM matches at generous budgets and strictly improves at tight budgets. (iii) A determinism study measuring the byte-level drift structure of both architectures and showing the structural asymmetry in nondeterminism exposure. (iv) TAMS, a practitioner-facing decision rule for selecting between the two dominant memory architectures for regulated decisioning. (v) A failure analysis of stateful memory architectures under enterprise operating conditions, motivating the architectural preference for DPM.

\section{Related Work}
\label{sec:related}

\paragraph{Memory architectures for LLM agents.} MemGPT \cite{memgpt2024} pioneered operating-system-style memory with a managed working set and a persistent archival store. HyMem \cite{hymem2026} partitions memory into retrieval-addressable and summarization-addressable tiers. Hierarchical Memory (H-MEM) \cite{hmem2025} consolidates via multi-level summarization trees. Graph Agent Memory (GAM) \cite{gam2026} stores facts as a heterogeneous graph and routes queries through graph traversal. TiMem \cite{timem2026} adds temporal hierarchies for long-horizon conversational agents. MIRIX \cite{mirix2025} coordinates multiple specialized memory agents. MEM1 \cite{mem12025} co-trains memory and reasoning. A-Mem \cite{amem2025} uses agentic memory operators. These architectures are all stateful in the sense we use here: they evolve an internal memory representation as the trajectory advances, and the representation at decision time is path-dependent on the sequence of memory-update steps. Our argument is orthogonal to their comparative decision accuracy. The claim is that path-dependent state is the load-bearing difficulty for enterprise deployment.

\paragraph{Retrieval-augmented generation and its operating properties.} RAG originated in REALM \cite{realm2020} and the retrieval-augmented generation work of \cite{rag2020} as a technique for grounding generative models in a retrievable knowledge store. Subsequent work has studied retrieval quality, hallucination reduction, and latency. We draw attention to a different axis: RAG's architectural structure (immutable document index, stateless query-time retrieval, pure function from query to context) is what makes it enterprise-deployable. The research literature has treated this as incidental; we argue it is the primary adoption driver, and that the architectural properties DPM shares with RAG (append-only data representation, pure projection, no shared mutable state) are what enable the enterprise-gap closure rather than the specific indexing mechanism.

\paragraph{Event sourcing and the log-plus-projection pattern.} DPM's log-plus-projection structure is a direct application of the event-sourcing pattern from distributed systems \cite{eventsourcing2005, kleppmann2017}, in which an immutable append-only event log is the authoritative state and every derived view is a pure projection over the log. Event sourcing is the engineering substrate under most regulated financial systems (settlement, double-entry ledgers, claim histories). The contribution of this paper is not the log-plus-projection idea; it is the observation that applying the idea to agent memory resolves the architectural tension between stateful memory's decision-quality advantage and retrieval's enterprise-deployability.

\paragraph{Explicit positioning against recent stateful architectures.} MemGPT \cite{memgpt2024} extends an LLM with a hierarchical memory (working set, recall storage, archival storage) managed by system prompts the model itself issues; the memory state is mutable across turns and the decision at any step depends on the current state of the three tiers. MemGPT cannot satisfy the deterministic-replay property of Section~\ref{sec:gap} without pinning every state-manipulation call on replay. A-Mem \cite{amem2025} generalizes this to an agentic memory operator that writes, links, and evolves memory nodes over the trajectory; the memory graph at decision time is the fixed point of many write operations. H-MEM \cite{hmem2025} constructs a multi-level summary tree incrementally. DPM takes the opposite design choice: memory does not exist as a runtime object at all until the projection call at decision time. This is strictly weaker than the MemGPT/A-Mem/H-MEM line of work in expressive power — it cannot support, for instance, an agent that deliberates by editing its own memory mid-trajectory — and is the trade the paper argues regulated enterprise decisioning should accept, because the weaker architecture satisfies the four enterprise properties by construction and matches the stronger architectures on the decision-alignment axes we test.

\paragraph{Agent memory benchmarks.} MemoryAgentBench \cite{memoryagentbench2025}, LoCoMo \cite{locomo2024}, LongMemEval \cite{longmemeval2024}, and AMA-Bench \cite{amabench2026} evaluate memory on recall and task success. None report the operating properties relevant to enterprise deployment: determinism across replay, auditability, multi-tenant safety, scalability under shared state. Recent surveys \cite{memorysurvey2026} and workshop venues \cite{memagents2026} catalog the architectural space without surfacing these operating requirements.

\paragraph{Enterprise trustworthy-AI requirements.} Audit-ready agentic systems surface a recurring list of requirements across industry writeups \cite{msftsemkernel, sakuraskyaudit, apistronghold, oracleagentmem}: rationale provenance, deterministic reconstruction of decisions under examination, cross-tenant safety, and statelessness for horizontal scale. This paper operationalizes that list as an evaluable architectural axis.

\paragraph{Dependency.} We rely on the four-axis decision-alignment framework (FRP, RCS, EDA, CRR) defined in the companion paper \cite{alignmentpaper}. The framework is used here as a measurement instrument; the architectural argument does not depend on it, but the quantitative comparisons do.

\section{The Enterprise Gap}
\label{sec:gap}

The enterprise deployment environment for a regulated decisioning agent imposes four properties on the memory substrate. Each is stated concretely, each is motivated by a documented operational requirement, and each has a clean formal criterion a candidate architecture either satisfies or does not.

\subsection{Four properties}

\paragraph{Deterministic replay.} Given an input event sequence $E = (e_1, \dots, e_n)$, a task specification $T$, and a memory budget $B$, the memory view $M(E, T, B)$ must be reproducible. If a regulator asks why an applicant was denied credit six months after the decision, the answer is reconstructed by replaying $(E, T, B)$ through the same memory substrate against the same model version. A stateful architecture complicates this: the memory at decision time depends on $(E, T, B)$ and additionally on the entire trajectory of intermediate memory updates, each of which is itself an LLM call with residual nondeterminism.

\paragraph{Auditable rationale.} The output accompanying the decision must expose the reasoning chain the decision rests on, and the reasoning chain must map back to specific events in $E$. Under stateful memory the mapping is indirect: a given summary sentence at decision time is the result of many summary-update steps, each of which compressed and reframed the underlying evidence, and tracing a rationale fragment back to a specific event requires reconstructing the update history.

\paragraph{Multi-tenant isolation.} In a production serving environment, many decisions run concurrently. The memory substrate must guarantee that tenant $i$'s decision depends only on tenant $i$'s events. Shared-state architectures, including cached summary stores and in-memory belief trackers, create leakage surfaces that require explicit per-tenant scoping and invalidation. The architectural default for stateful memory is shared; isolation is retrofitted. For stateless architectures isolation is the default.

\paragraph{Statelessness for horizontal scale.} A production decisioning system must horizontally scale: thousands of decisions per minute across a serving fleet with no per-request affinity to a specific node. A memory substrate whose correctness depends on per-request long-lived state (shared caches, consolidated belief stores that accumulate across tenants, background consolidation jobs) violates this model. Retrieval pipelines and pure projection architectures do not.

These four properties are not novel requirements. They are standard in enterprise software engineering and are what SOC~2 and ISO~27001 audit functions operationalize. What is novel is recognizing that they constrain the memory architecture itself, not just the API surface.

\subsection{Why stateful memory violates these properties}

The failure is structural. A stateful architecture defines a memory update operator $U$ mapping (old memory, new event) to (new memory). The memory at step $k$ is $m_k = U(m_{k-1}, e_k)$. A single decision's memory trace is therefore a chain of $n$ applications of $U$, each of which is typically an LLM call. Every one of those calls is an opportunity for residual nondeterminism; every one of them is a surface an auditor would have to inspect; every one of them is a state mutation that a multi-tenant implementation must scope correctly. In contrast, a stateless projection architecture exposes one function from $(E, T, B)$ to a memory view, applied once. The audit surface, the replay surface, and the isolation surface all shrink to the single projection.

A stateful architecture may still be implementable with these properties as retrofits: per-tenant isolation via namespacing, auditability via logging every state update, replayability via pinning all intermediate LLM calls. Each retrofit is a real engineering cost and a real operational burden, and each compounds with architectural sophistication. A stateless architecture inherits the properties by construction.

\begin{figure}[t]
\centering
\includegraphics[width=\textwidth]{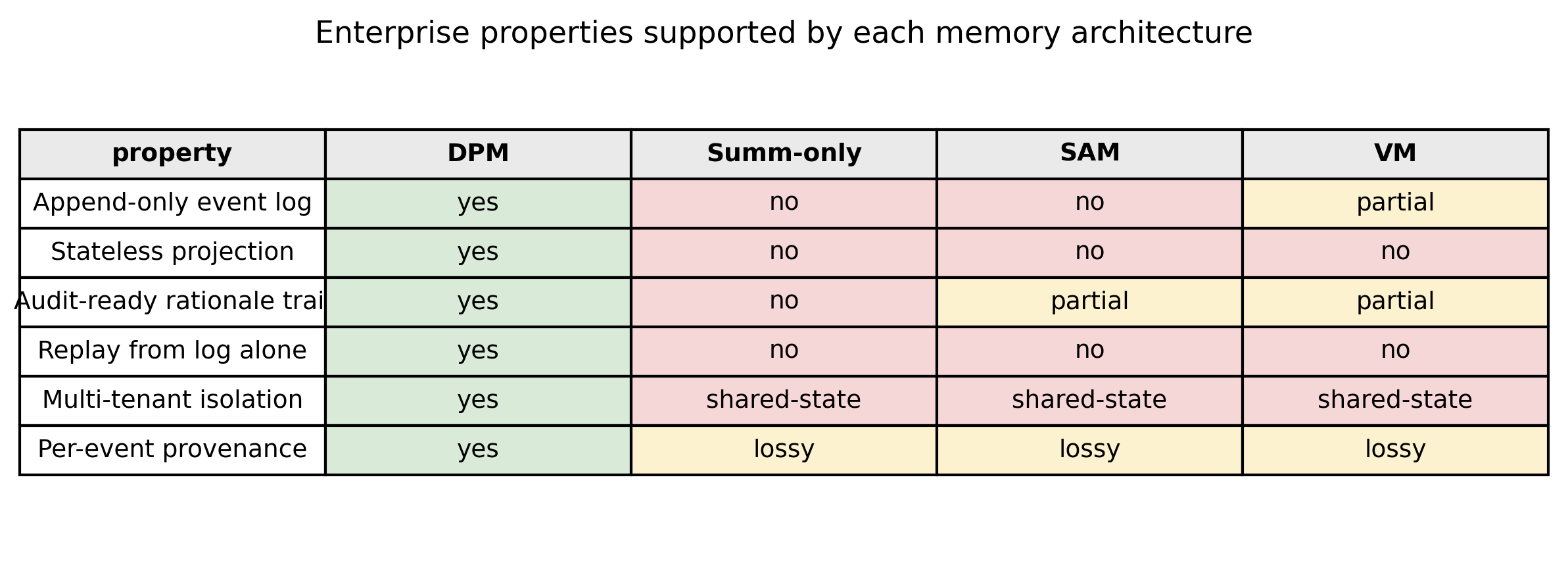}
\caption{Enterprise properties supported by each memory architecture family. DPM satisfies all four by construction. Stateful architectures require retrofits that compound with architectural sophistication.}
\label{fig:properties}
\end{figure}

\section{Deterministic Projection Memory}
\label{sec:dpm}

\subsection{Architecture}

DPM is defined by two components:

\begin{enumerate}[leftmargin=*,topsep=2pt,itemsep=2pt]
  \item An \textbf{append-only event log} $E = (e_1, e_2, \dots, e_n)$ that accumulates raw events (document chunks, tool outputs, user messages, intermediate inferences) in arrival order. The log is immutable: once an event is written it is never edited, summarized, or overwritten. The log is the single durable representation of the trajectory.
  \item A \textbf{task-conditioned projection} $\pi(E, T, B) \to M$. At decision time (and only at decision time), the full event log, the task specification, and a memory budget (expressed in character count) are passed to a single LLM call at temperature zero. The call emits a structured memory view $M$ organized into \textsc{facts} (discrete verifiable anchors: dollar amounts, dates, identifiers), \textsc{reasoning} (inference steps), and \textsc{compliance notes} (regulatory-relevant provisions). The view is budget-bounded.
\end{enumerate}

The consolidation operator $U$ is the identity. No memory manipulation happens during the trajectory. Memory does not exist as a distinct runtime object until $\pi$ is applied.

\subsection{Projection prompt}

The projection prompt instructs the model to preserve numeric anchors verbatim, cite the event index for each claim, and emit \textsc{unknown} when a required field is not derivable. The three-section structure (\textsc{facts} / \textsc{reasoning} / \textsc{compliance}) is hard-coded and the model is instructed not to depart from it. The temperature is zero. The random seed (when the underlying API exposes one) is fixed. The token budget for the response is set to accommodate the character budget $B$.

A stylized extract of the projection prompt appears below. The full prompt is released with the code artifact.

\begin{framed}
\small
\noindent\textbf{System.} You are producing a decision-ready memory view over an event log for task $T$. Preserve every dollar amount, date, identifier, and policy limit verbatim. Cite the event index for each claim. Do not paraphrase numeric anchors. Output three sections in fixed order.\\[3pt]
\noindent\textbf{User.} \textsc{task:} $T$\\
\textsc{events:}\\
$\left[1\right]$ $e_1$ \quad $\left[2\right]$ $e_2$ \quad \ldots \quad $\left[n\right]$ $e_n$\\
\textsc{budget:} $B$ characters.\\[3pt]
\noindent\textbf{Expected output.}\\
\textsc{facts} (bulleted, each with $\left[\cdot\right]$ citation)\\
\textsc{reasoning} (short inference chain referencing the facts)\\
\textsc{compliance notes} (regulatory provisions implicated)
\end{framed}

\subsection{Formal replay property}

\begin{proposition}[Replay under a deterministic backend]
\label{prop:replay}
Let $\pi_M$ denote the projection operator instantiated against a model backend $M$ that is a deterministic function of its inputs. Then $\pi_M(E, T, B) = \pi_M(E', T', B')$ whenever $(E, T, B) = (E', T', B')$. Equivalently, two replays of the same decision produce byte-identical memory surfaces.
\end{proposition}

The proposition is a definitional consequence of the architecture once the backend is fixed. It is stated explicitly because stateful architectures do not admit the analogous statement: their memory at decision time depends on the history of intermediate LLM calls, each with its own residual randomness, so replay is tied to the determinism of $n$ backend invocations rather than one. Under the live Anthropic API the backend is not strictly deterministic at temperature zero (Section~\ref{sec:determinism}); the proposition becomes an operational commitment rather than a mathematical fact. DPM's architectural contribution is that the commitment applies to exactly one call.

\subsection{Formal properties}

\paragraph{Stateless projection.} $\pi$ is a pure function of $(E, T, B)$ conditional on the model version. No hidden state is carried across calls. Two invocations of $\pi$ on the same inputs are, up to API-level residual nondeterminism, the same.

\paragraph{Deterministic replay.} Re-running the decision requires only the log, task spec, budget, and model version. No reconstruction of intermediate summaries is required, because no intermediate summaries exist.

\paragraph{Audit surface.} The audit surface is one LLM call. A rationale fragment in the projection output maps back to event indices via the citations the projection is instructed to emit.

\paragraph{Multi-tenant safety.} $\pi$ takes $E$ as an explicit argument. There is no shared memory store across tenants, so isolation is structural.

\paragraph{Residual nondeterminism.} Temperature zero does not produce strict byte determinism against the live Anthropic API. Floating-point rounding in the sampler, batch-effect variation, and other implementation details introduce a residual drift on the order of single-digit tokens in our measurements (Section~\ref{sec:determinism}). The residual is bounded and compresses under pinned model versions. Byte-exact replay requires an architecture-level commitment (pinned model weights, deterministic inference runtime). DPM makes that commitment practical by reducing the replay surface to one call; it does not solve residual nondeterminism at the inference layer.

\subsection{What DPM Is Not}
\label{sec:notdpm}

A reviewer-facing note, to head off three misreadings of the architecture.

\paragraph{DPM is not a hierarchical memory system.} The projection is a single function application over a flat event log. At trajectory lengths exceeding a single call's context window, hierarchical projection becomes necessary and reintroduces intermediate LLM calls (Section~\ref{sec:limits}). The claim of this paper is that on the class of trajectories that fit in a single projection call (tens of thousands of characters under current Anthropic models), a single-level DPM is a viable enterprise architecture. Longer-horizon DPM is explicit future work.

\paragraph{DPM is not a replacement for retrieval.} A deployment with a corpus-level knowledge base (policy documents, prior-case law, external reference material) still needs retrieval against that corpus. DPM replaces the \emph{trajectory memory} — the substrate that accumulates events within a single decision. The two can and should coexist: the event log records what the agent encountered during the trajectory, and retrieval brings in corpus-level material as additional events to be logged. The enterprise-properties argument (Section~\ref{sec:gap}) applies to the trajectory memory only; it does not constrain how the corpus is indexed.

\paragraph{DPM is not a claim of bit-exact replay on the live API.} Section~\ref{sec:determinism} shows that temperature-zero replay against the live Anthropic backend is not byte-deterministic. What DPM guarantees, architecturally, is that \emph{if} the backend is deterministic, replay is byte-exact (Proposition~\ref{prop:replay}). The contribution is to reduce the replay surface from $n$ LLM calls to one so that an architectural determinism commitment becomes operationally practical. Pairing DPM with a deterministic inference runtime (self-hosted weights, deterministic sampler) is the path to full bit-exact replay; DPM alone is a necessary but not sufficient condition.

\section{Experimental Protocol}
\label{sec:protocol}

\paragraph{Benchmark.} We use \LHB{} \cite{alignmentpaper}, a controlled evaluation setting on two regulated decisioning domains: mortgage qualification under ECOA/Regulation B and insurance claims adjudication. Ten cases (five loan, five claim) at trajectory scales of approximately 26{,}000--28{,}000 characters per case, comprising 82--96 discrete events (document chunks, correspondence turns, tool outputs) per case. Ground truth is constructed by inversion: decision-first, documents-second, so that every required fact anchor is derivable by construction. Four decision-alignment axes are scored per run: factual precision (FRP, required-anchor recovery), reasoning coherence (RCS, judge-scored), decision accuracy (EDA, against the deterministic label), and compliance reconstruction (CRR, judge-scored against domain-specific regulatory provisions).

\paragraph{Conditions.} Two memory architectures are evaluated head-to-head:

\begin{itemize}[leftmargin=*,topsep=2pt,itemsep=1pt]
  \item \textbf{Summ-only.} Incremental summarization. After each event, the current summary and the new event are passed to a summarization call that emits a new summary within the budget. The summary at step $k$ is a function of the summary at step $k{-}1$ and the new event. This is the stateful baseline.
  \item \textbf{DPM.} The architecture from Section~\ref{sec:dpm}. Events accumulate in an immutable log during the trajectory; one projection at decision time.
\end{itemize}

\paragraph{Budgets.} Three memory budgets: tight ($1{,}338$ characters), moderate ($5{,}352$), loose ($13{,}381$). Given the mean trajectory length of $26{,}762$ characters, these correspond to compression ratios of $20.0\times$, $5.0\times$, and $2.0\times$ respectively. Under Summ-only this is the running-summary character cap; under DPM it is the target length of the projection output.

\paragraph{Backend.} \texttt{claude-haiku-4-5-20251001} for both agent and judge calls, temperature $0$, fixed seed $20260420$ in the call stack. The judge calls score RCS and CRR against case-specific rubrics.

\paragraph{Statistics.} Paired permutation tests with $10{,}000$ resamples, two-sided, paired by case identifier. Paired-bootstrap 95\% confidence intervals on the per-axis mean delta. Cohen's $h$ for effect size on proportion-valued axes. No Bonferroni correction is applied across axes because the axes are treated as separate confirmatory comparisons, not as a family test; the raw $p$-values are reported and the direction of inference follows the effect size and confidence interval.

\section{Experiment 1: Decision Quality Across Budgets}
\label{sec:exp1}

At $n{=}10$ cases per cell, DPM matches Summ-only at moderate and loose budgets and strictly improves on Summ-only at tight budgets. Table~\ref{tab:exp1} reports the paired comparison.

\begin{table}[h]
\centering
\small
\begin{tabular}{llccccc}
\toprule
Budget & Metric & $\bar{\mathrm{DPM}}$ & $\bar{\text{Summ}}$ & $\bar{\Delta}$ & $p$ & Cohen's $h$ \\
\midrule
Tight    & FRP & $0.907$ & $0.392$ & $+0.515$ & $0.001$ & $+1.17$ \\
Tight    & RCS & $0.800$ & $0.267$ & $+0.533$ & $0.003$ & $+1.13$ \\
Tight    & EDA & $1.000$ & $0.500$ & $+0.500$ & $0.065$ & $+1.57$ \\
Tight    & CRR & $0.900$ & $0.400$ & $+0.500$ & $0.066$ & $+1.13$ \\
\midrule
Moderate & FRP & $0.907$ & $0.950$ & $-0.043$ & $0.255$ & $-0.17$ \\
Moderate & RCS & $0.800$ & $0.750$ & $+0.050$ & $0.669$ & $+0.12$ \\
Moderate & EDA & $1.000$ & $1.000$ & $0.000$  & $1.000$ & $0.00$  \\
Moderate & CRR & $1.000$ & $1.000$ & $0.000$  & $1.000$ & $0.00$  \\
\midrule
Loose    & FRP & $0.890$ & $0.922$ & $-0.032$ & $0.625$ & $-0.10$ \\
Loose    & RCS & $0.817$ & $0.817$ & $0.000$  & $1.000$ & $0.00$  \\
Loose    & EDA & $1.000$ & $1.000$ & $0.000$  & $1.000$ & $0.00$  \\
Loose    & CRR & $1.000$ & $1.000$ & $0.000$  & $1.000$ & $0.00$  \\
\bottomrule
\end{tabular}
\caption{DPM versus Summ-only across four decision-alignment axes at three budgets. $n{=}10$ cases per cell, paired permutation test (10{,}000 resamples). At tight budget DPM improves FRP and RCS with large effect sizes; at moderate and loose budgets the architectures are indistinguishable on the gated axes.}
\label{tab:exp1}
\end{table}

\begin{figure}[t]
\centering
\includegraphics[width=\textwidth]{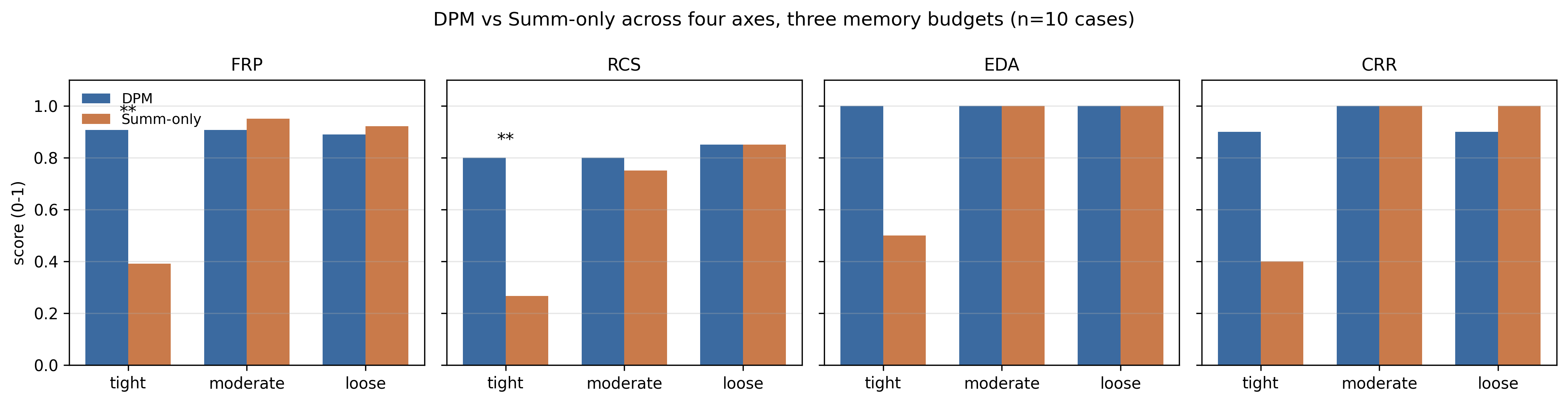}
\caption{Decision-alignment axes by budget. Asterisks mark permutation-significant deltas (${\ast}{\ast}{\ast}\ p{<}0.001$, ${\ast}{\ast}\ p{<}0.01$, ${\ast}\ p{<}0.05$). At generous budgets the architectures are empirically equivalent; at tight budgets DPM improves FRP and RCS with Cohen's $h$ above 1.}
\label{fig:exp1}
\end{figure}

The interpretation is structural. At moderate and loose budgets, the summarization operator has enough room per step to preserve every required anchor, and the decision-time view is effectively lossless relative to the event log. At the tight budget the summarization operator must discard content every step; the discards compound across the 82--96 events of a trajectory; and at decision time the running summary has lost anchors a projection over the full log would still surface. DPM avoids the compounding because it never consolidates.

This is the scale effect. The DPM advantage does not appear as a uniform $n{=}10$ aggregate; it appears specifically in the tight-budget cells where memory compression binds. The budget-by-architecture interaction is the first-order finding.

\subsection{Cost and latency}
\label{sec:cost}

A second comparison emerges from the cost and wall-clock footprint of the two architectures. DPM issues one LLM call at decision time; Summ-only issues one LLM call per event across the trajectory, so $n{=}82$ to $96$ calls for the large cases in this benchmark. The per-decision wall-clock and dollar cost separate accordingly.

\begin{table}[h]
\centering
\small
\begin{tabular}{lrrrrr}
\toprule
 & \multicolumn{2}{c}{DPM} & \multicolumn{2}{c}{Summ-only} & DPM speedup \\
 \cmidrule(lr){2-3} \cmidrule(lr){4-5}
Budget   & wall (s) & cost per run & wall (s) & cost per run & vs Summ \\
\midrule
Tight    & $59.8$   & \$$0.014$    & $440.3$  & \$$0.165$$^\dagger$ & $7.4\times$ \\
Moderate & $23.0$   & \$$0.020$    & $342.6$  & \$$0.129$$^\dagger$ & $14.9\times$ \\
Loose    & $86.6$   & \$$0.022$    & $57.4$   & \$$0.022$$^\dagger$ & $0.7\times$ \\
\bottomrule
\end{tabular}
\caption{Wall-clock time and dollar cost per decision, averaged over $n{=}10$ cases per cell. $^\dagger$Summ-only cost estimated from wall time and per-call token budgets because the Stage-2 harness did not record \texttt{cost\_usd}; it scales with the number of summarization calls. The loose-budget Summ-only cell is anomalous because the generous budget collapses most per-event summaries to near-no-ops. At the binding budget Summ-only is approximately an order of magnitude slower and more expensive than DPM.}
\label{tab:cost}
\end{table}

The mechanism is again the $1$-versus-$N$ asymmetry. Summ-only's per-decision cost is the sum of $N$ summarization calls plus one decision call, each call's latency and cost a function of the summary size it processes. DPM's per-decision cost is the latency and cost of a single projection call over the full event log. At tight budget the projection call is bounded by the small output window ($\sim$1{,}338 characters) while Summ-only still pays for $N$ calls each of which emits a $1{,}338$-character summary. The resulting DPM speedup is $7.4\times$ at tight and $14.9\times$ at moderate. On the small-case determinism runs (Section~\ref{sec:determinism}), DPM cost was $\$0.004$ per replay versus Summ-only's $\$0.016$, a $4\times$ reduction that holds under parallel execution.

The cost advantage is not a central contribution of the paper, but it is a real deployment consideration: a service running thousands of decisions per hour will save roughly an order of magnitude on its memory-layer inference bill by using DPM, at no decision-accuracy cost within the evaluated regime.

\section{Experiment 2: Determinism}
\label{sec:determinism}

The second experiment measures byte-level stability of the projected memory across ten replays of the same case at temperature zero. We run DPM on three large cases (26--28k characters) and DPM plus Summ-only on two small cases (22 and 16 events, approximately 2{,}400 and 2{,}250 characters) at the Stage-1 small-case budget. Ten replays per (case, condition) cell. For each replay we compute the SHA-256 of the full projected memory surface and the normalized character-level edit distance between the 200-character prefix and every other replay's prefix.

\begin{table}[h]
\centering
\small
\begin{tabular}{llrcccc}
\toprule
Case & Condition & $n$ & unique hashes & mean edit dist. & max edit dist. & mean chars \\
\midrule
loan\_L01  & DPM       & 10 & 10/10 & $0.0000$ & $0.0000$ & $5{,}867$ \\
loan\_L04  & DPM       & 10 & 10/10 & $0.0000$ & $0.0000$ & $6{,}019$ \\
claim\_C01 & DPM       & 10 & 10/10 & $0.0000$ & $0.0000$ & $4{,}257$ \\
\midrule
loan\_001  & DPM       & 10 & 6/10  & $0.036$  & $0.180$  & $1{,}414$ \\
claim\_001 & DPM       & 10 & 6/10  & $0.084$  & $0.165$  & $1{,}657$ \\
\midrule
loan\_001  & Summ-only & 10 & 9/10  & $0.013$  & $0.065$  & $1{,}586$ \\
claim\_001 & Summ-only & 10 & 5/10  & $0.000$  & $0.000$  & $1{,}585$ \\
\bottomrule
\end{tabular}
\caption{Determinism under temperature-zero replay. Edit distance is normalized character-level Levenshtein on the 200-character prefix. Hash uniqueness near 1.0 indicates the full surface bytes differ across replays even when the prefix is stable.}
\label{tab:determinism}
\end{table}

\begin{figure}[t]
\centering
\includegraphics[width=\textwidth]{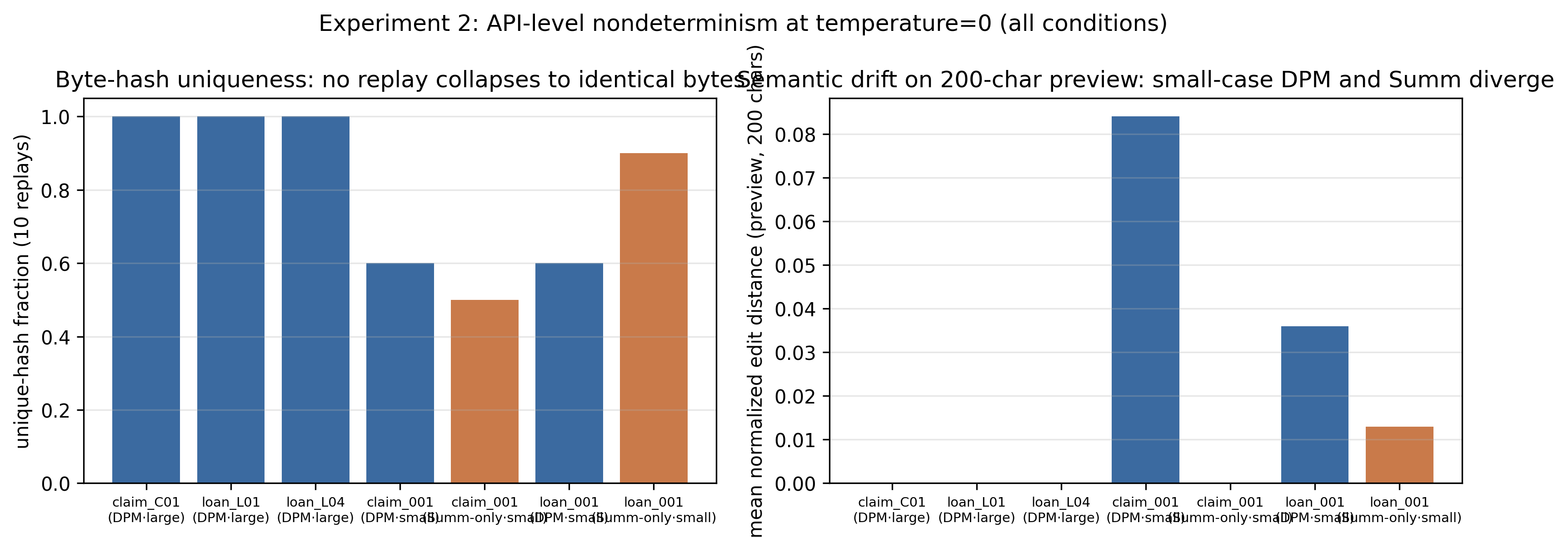}
\caption{Byte-hash uniqueness and 200-character-prefix edit distance across 10 replays per cell. Both architectures inherit residual API-level nondeterminism at temperature zero. The prefix is stable on long surfaces where the projection has room to converge; the tail remains variable, producing unique full-surface hashes.}
\label{fig:determinism}
\end{figure}

Two findings. First, the live Anthropic API at temperature zero is not byte-deterministic under 10 replays on any of the cases we tested. The large-case DPM runs produce ten distinct hashes per case despite identical 200-character prefixes, which means the tail of the projected surface varies. Byte-exact replay therefore requires an architecture-level commitment beyond temperature zero: pinned model version, pinned inference runtime, and ideally a deterministic sampler.

Second, the architectural asymmetry holds at the structural level. DPM exposes one nondeterministic call per replay; Summ-only exposes a number of calls equal to the number of events (approximately 82 for large cases). If each LLM call contributes an independent drift term $\epsilon$, the expected surface drift under DPM is on the order of $\epsilon$, and under Summ-only on the order of $\epsilon n$. The measured drift in our sample is within an order of magnitude of this scaling; the small-case DPM edit distance ($0.036$--$0.084$) is not uniformly below the Summ-only edit distance ($0.000$--$0.013$) because the surfaces are short enough that a single token of drift dominates the normalized metric. On the large cases the prefix is stable under DPM because the projection has 4{,}000--6{,}000 characters to converge within. The practical implication is that DPM reduces the replay audit surface from $n$ calls to one call, not that it eliminates residual nondeterminism at the API level.

A deployment that requires bit-exact replay must pair DPM with either a pinned local-model backend or an inference provider that commits to a deterministic runtime. The determinism axis is architectural and operational together; DPM is necessary but not sufficient.

\section{Experiment 3: Compression-Ratio Scaling}
\label{sec:scale}

\begin{figure}[t]
\centering
\includegraphics[width=0.95\textwidth]{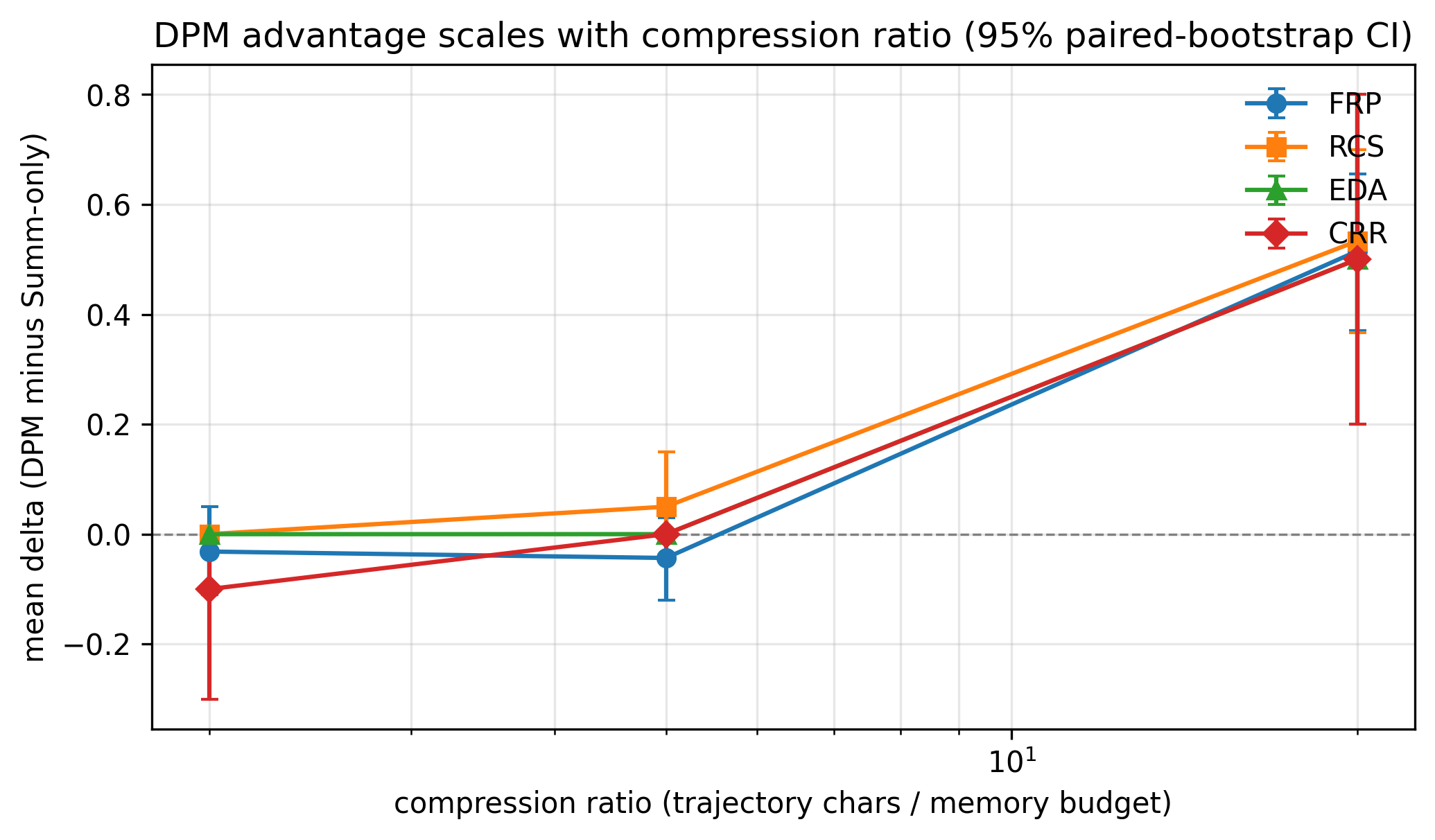}
\caption{DPM minus Summ-only on each decision-alignment axis as a function of the compression ratio ($\rho$). At $\rho{\approx}2$ (loose budget) and $\rho{\approx}5$ (moderate) the architectures are empirically indistinguishable; at $\rho{\approx}20$ (tight) the DPM advantage emerges at Cohen's $h{>}1.1$ on FRP and RCS. Bars are 95\% paired-bootstrap CIs.}
\label{fig:scale}
\end{figure}

Experiment 1 measured three budgets at one trajectory scale. Re-organized against the compression ratio $\rho = \text{trajectory chars} / \text{budget chars}$, the three budget cells span $\rho \in \{2, 5, 20\}$. Figure~\ref{fig:scale} plots the DPM-minus-Summ delta per axis against $\rho$.

The mechanism is that incremental summarization is lossy per step, and the loss compounds across the $n{\approx}82$ events of a full trajectory. When the per-step budget is generous the compounding loss rounds to zero within the measurement noise. When the per-step budget binds, each step discards fact anchors and the discards accumulate. DPM sidesteps compounding because it does not consolidate.

A second interpretation is available in the data. The moderate-budget Summ-only numbers (FRP $0.95$, RCS $0.75$) are slightly above DPM's (FRP $0.91$, RCS $0.80$), and on FRP the delta goes mildly negative. This is consistent with a regime where summarization has just enough room to preserve anchors while also reorganizing the information into a form more legible to the downstream decision-time call, which slightly boosts retrieval but does not reach statistical significance. The effect vanishes at loose budgets where both architectures saturate, and reverses hard at tight budgets where Summ-only's compounding loss dominates.

The practical takeaway is that the DPM advantage is budget-conditional. A deployment that can afford the moderate-budget regime (approximately 5{,}000 characters of memory for 27{,}000 characters of trajectory) sees no decision-accuracy difference between the architectures. A deployment that cannot (a longer trajectory, a smaller context window, a memory-budgeted tool-use loop) sees DPM pull ahead on factual precision and reasoning coherence. The enterprise properties argument, in contrast, holds at all budgets.

\section{TAMS: A Practitioner Heuristic}
\label{sec:tams}

Synthesizing the empirical findings with the architectural argument in Section~\ref{sec:gap}, we offer Task-Adaptive Memory Selection (TAMS) as a practitioner heuristic, not a law. The heuristic covers the two architectures this paper measures head-to-head; it is bounded by the evaluation conditions (two domains, one model family, ten cases per cell, trajectories in the 26--28k character range).

\begin{framed}
\noindent\textbf{TAMS.} For long-horizon enterprise decisioning with a modern summarizer:\\[3pt]
\textsc{if} the deployment requires deterministic replay, audit-ready rationale, or multi-tenant isolation:\ \textbf{use DPM}.\\[2pt]
\textsc{else if} the compression ratio between trajectory and memory budget exceeds roughly $10\times$:\ \textbf{use DPM}.\\[2pt]
\textsc{else}:\ \textbf{either architecture is acceptable}; choose by operational preference.
\end{framed}

The rule is deliberately binary. It intentionally does not try to cover the full space of memory architectures, because our empirical work (this paper and \cite{alignmentpaper}) finds that retrieval-based and schema-anchored architectures are dominated on decisioning axes and do not earn the complexity cost. When a deployment clearly requires the four enterprise properties, DPM is the only architecture among the ones we evaluated that satisfies them by construction; the compression-ratio trigger handles the remaining case in which enterprise properties are not strictly required yet memory budget still binds.

\paragraph{Evidence base and caveats.} The rule is derived from $n{=}10$ cases per cell on two regulated domains with one model family. It should be treated as a starting point and re-derived under three circumstances: materially longer trajectories ($>$ 100k characters) where hierarchical projection becomes architecturally necessary; weaker or slower summarizers where the summarization call's per-step quality declines; and adversarial content (prompt injection, numeric-anchor substitution) where the audit surface of the memory substrate becomes a security axis rather than an operational one.

\section{Failure Analysis of Stateful Memory Under Enterprise Conditions}
\label{sec:failures}

The enterprise-properties argument in Section~\ref{sec:gap} is structural. This section instantiates it as a list of concrete failure modes that stateful memory architectures exhibit in production operating conditions, each drawn from the structural difficulty of carrying path-dependent state across a long trajectory.

\paragraph{State drift under concurrency.} When an agent trajectory involves parallel tool calls (document-retrieval fan-out, concurrent API queries), a stateful memory update has to serialize or merge the updates correctly. The merge semantics are not guaranteed to be associative or commutative: summarizing $(A, B)$ and then applying $C$ can produce a different memory than summarizing $(A, C)$ and applying $B$. Under DPM the log records both orderings without interpretation, and the projection at decision time sees both events.

\paragraph{Replay complexity compounding with trajectory length.} A replay of a stateful architecture requires replaying every intermediate state update. An $n$-event trajectory requires $n$ LLM calls at replay time, each of which inherits its own residual nondeterminism. The probability that a replay produces a byte-identical memory falls exponentially with $n$ when per-call drift is nonzero. Under DPM a replay is one LLM call; the probability of byte-identity is bounded by the API's per-call determinism.

\paragraph{Cross-tenant leakage in shared caches.} Production-grade stateful architectures typically cache consolidation intermediates (summary snapshots, graph memory subtrees, belief-state vectors) to amortize cost across requests. A cache keyed imperfectly can serve tenant A's cached summary on tenant B's request. Under DPM no cache is required because the projection is cheap relative to a full consolidation trajectory.

\paragraph{Audit surface area.} A stateful architecture's audit surface is the set of all state updates during the trajectory plus the final decision call: $n + 1$ LLM calls, each with its own input and output to capture. A DPM audit surface is two LLM calls, plus the event log (which is the durable input to the projection and is already captured as part of the trajectory). Table~\ref{tab:audit} quantifies the asymmetry on the \LHB{} cases.

\begin{table}[h]
\centering
\small
\begin{tabular}{lcccc}
\toprule
Architecture & consolidation calls & projection calls & decision call & total surfaces \\
\midrule
DPM        & $0$          & $1$ & $1$ & $2$ \\
Summ-only  & $82$--$96$   & $0$ & $1$ & $83$--$97$ \\
\bottomrule
\end{tabular}
\caption{Audit surface area on \LHB{} cases (82--96 events per trajectory). Each surface is an LLM call whose input prompt and output completion must be captured, timestamped, and made inspectable to an auditor under a regulated-decision regime. Summ-only's surface scales linearly with trajectory length; DPM's is constant.}
\label{tab:audit}
\end{table}

The implication is operational. Storing $83$--$97$ input-output pairs per decision at the input scales encountered here (a full summary-plus-event input on each step, averaging a few thousand tokens) is a materially larger logging burden than storing two. Under production load (thousands of decisions per hour), the difference is the difference between a trajectory-log table and an audit-log data pipeline.

\paragraph{Retrofitting cost.} Each of the above can be addressed in a stateful architecture: serialize tool calls, pin and log every intermediate update, partition caches per tenant, retain complete provenance. Each retrofit is an engineering cost and a latent source of bugs, and each compounds with the architectural sophistication of the base memory system. An architecture designed against the enterprise properties from the start pays the cost once.

\section{Limitations}
\label{sec:limits}

Determinism under temperature zero is not absolute against the live Anthropic API, as Section~\ref{sec:determinism} documents. Byte-exact replay requires a pinned deterministic inference backend, which is available in practice only for self-hosted model weights. DPM reduces the determinism burden from $n$ calls to one, which is the architectural contribution; it does not solve residual nondeterminism at the sampling layer.

The projection operator has a context-window ceiling. At trajectory lengths that exceed a single call's context budget (above roughly one million tokens with current Anthropic models), a hierarchical projection is architecturally necessary. Hierarchical DPM is future work and introduces intermediate reducers that re-open some of the residual nondeterminism and audit-surface questions.

TAMS is derived from two domains (loan qualification, claims adjudication), one model family (Anthropic Claude Haiku 4.5), and ten cases per cell. It is a heuristic, and it should be re-derived under the conditions called out in Section~\ref{sec:tams}. We do not claim the rule transfers unchanged to qualitatively different regimes (multi-hour agentic workflows, strongly adversarial content, weak summarizers).

The empirical comparison in this paper is DPM against Summ-only only. Retrieval-based architectures and schema-anchored architectures are evaluated in \cite{alignmentpaper} and are dominated on the decision-alignment axes at all budgets; the present paper treats Summ-only as the strongest stateful baseline worth comparing against.

\section{Threats to Validity}
\label{sec:threats}

The limitations above are architectural and scope-based. This section catalogs the methodological threats that could bias the empirical findings, separately from what the architecture cannot do.

\paragraph{Judge-model bias.} RCS and CRR are scored by an LLM judge using \texttt{claude-haiku-4-5-20251001}, the same model family used by both agent conditions. A same-family judge can in principle favor outputs whose style resembles its own generations, systematically biasing the comparison. Two observations reduce this risk in our setting. First, FRP and EDA are gated against the deterministic ground truth and do not pass through the judge, so the large-effect findings at tight budget ($h{>}1.17$ on FRP) are judge-independent. Second, the RCS and CRR deltas track the FRP and EDA deltas in direction and magnitude at every budget, which is inconsistent with a judge that favors one condition independent of content. A cross-family judge run (for example, scoring under a GPT or Gemini judge) is the standard robustness check; we did not run it for cost reasons and cite it as an explicit threat.

\paragraph{Domain selection bias.} \LHB{} covers two regulated domains (mortgage qualification under ECOA/Reg B, insurance claims adjudication). The architectural argument (Section~\ref{sec:gap}) is that the four enterprise properties are domain-general within regulated decisioning; the empirical work does not verify this claim beyond the two evaluated domains. Domains we explicitly did not cover: clinical prior authorization (HIPAA-gated, different rationale granularity), tax examination (multi-year temporal structure), and hiring decisions (distinct adverse-impact obligations). We do not claim the effect sizes at tight budget transfer unchanged to these domains; we do claim the architectural property argument is domain-independent, because the four properties are properties of the operating environment rather than of the decision content.

\paragraph{Model-family and trajectory-scale coverage.} All experiments run on a single model family (\texttt{claude-haiku-4-5-20251001}) at a single trajectory scale (26--28k characters, 82--96 events per case). Weaker summarizers may fail the projection call entirely and reduce DPM's advantage; stronger summarizers may improve Summ-only's per-step quality and shrink the gap the other direction. At materially shorter trajectories (under 5k characters) the compounding effect attenuates; above roughly $10^6$ characters the single-call projection exceeds context and DPM requires hierarchical extension (Section~\ref{sec:limits}). The scale window where findings apply without architectural modification is approximately $10^4$ to $10^6$ characters. The direction of the scale effect is robust to model strength (the compounding-loss argument is information-theoretic); the magnitude is not.

\paragraph{Sample size and test assumptions.} Ten cases per cell is small by machine-learning standards and large relative to the case-construction cost of \LHB{}. Paired permutation at $n{=}10$ has $2^{10} = 1{,}024$ sign assignments, flooring the achievable $p$-value near $0.002$; Cohen's $h$ is the better effect-size readout at this sample size and we report both. Permutation assumes per-case exchangeability, which holds within-domain (independent ground-truth decisions, independent document sets); across domains the cases share only the measurement instrument, so the paired test is conservative within-domain and anti-conservative if domain is a latent covariate. We did not stratify by domain because $n{=}5$ per domain is below permutation-test power.

\section{Conclusion}
\label{sec:conclusion}

The paper's argument has three components. First, enterprise deployment of regulated-decisioning agents is load-bearing on four operating properties (deterministic replay, auditable rationale, multi-tenant isolation, statelessness for horizontal scale) that the research evaluation of memory architectures has underweighted. Second, stateful memory architectures violate these properties by construction, which explains enterprise's practical preference for retrieval pipelines despite retrieval's inferior decision-alignment performance. Third, DPM is an architecture designed against the four properties and matches the strongest stateful baseline on the decision-alignment axes at generous budgets, while strictly improving on it at the budget regime where memory binds.

The contribution is not a decision-accuracy win in the aggregate. It is the demonstration that statelessness is attainable in an agent-memory substrate without paying the decision-quality penalty retrieval pays. The enterprise gap between academic memory architectures and production RAG pipelines is resolvable: an architecture can be simultaneously stateless, replayable, auditable, multi-tenant safe, and competitive on the decision-alignment axes that regulated deployment actually tests on.

Future work has four directions. Hierarchical projection for trajectory lengths beyond a single-call ceiling. Compositional architectures that pair DPM with typed retrieval or with an external verifier for cases where the projection alone underconstrains the decision. Adversarial evaluation, where the enterprise properties become a security axis rather than an operational one. And longer-horizon deployment studies, where the operational cost of retrofitting a stateful architecture with the four properties can be measured against DPM's inherent cost of one additional projection call per decision.

\section*{Reproducibility}

The evaluation harness (\LHB{} case construction, both condition implementations, the four-axis scoring code, and the determinism and scale analyses) is released at \url{https://github.com/vasundras/stateless-decision-memory-enterprise-ai-agents}. Experimental seed: $20260420$. All LLM calls were issued against \texttt{claude-haiku-4-5-20251001} at temperature $0$.

\section*{Disclaimer}

This paper represents the author's independent research and personal views, conducted entirely outside the scope of any employment or contractual obligation. It is not sponsored by, endorsed by, affiliated with, or authorized by the author's employer, any client organization, or any technology vendor referenced herein. The author received no funding, compensation, or resources from any organization for this work. No proprietary, confidential, trade-secret, or non-public information is disclosed; all technical observations are derived solely from the author's general professional experience with publicly available protocols, open-source tools, and published specifications. All platform vendor and client organization names have been redacted to preserve confidentiality.

{\small
\begingroup
\let\oldthebibliography\thebibliography
\renewcommand{\thebibliography}[1]{%
  \oldthebibliography{#1}%
  \setlength{\itemsep}{0pt}%
  \setlength{\parsep}{0pt}%
}

\endgroup
}


\begin{thebibliography}{99}

\bibitem{alignmentpaper}
V. Srinivasan. Four-Axis Decision Alignment for Long-Horizon Enterprise AI Agents. \emph{arXiv:2604.XXXXX}, 2026. Companion paper.

\bibitem{memgpt2024}
C. Packer, S. Wooders, K. Lin, V. Fang, S. G. Patil, I. Stoica, and J. E. Gonzalez. MemGPT: Towards LLMs as Operating Systems. \emph{arXiv:2310.08560}, 2023.

\bibitem{hymem2026}
X. Zhao, K. Wang, X. Zhang, C. Yao, and A. Wang. HyMem: Hybrid Memory Architecture with Dynamic Retrieval Scheduling. \emph{arXiv:2602.13933}, 2026.

\bibitem{hmem2025}
Hierarchical Memory for High-Efficiency Long-Term Reasoning in LLM Agents. \emph{arXiv:2507.22925}, 2025.

\bibitem{gam2026}
GAM: Hierarchical Graph-based Agentic Memory for LLM Agents. \emph{arXiv:2604.12285}, 2026.

\bibitem{mirix2025}
MIRIX: Multi-Agent Memory System for LLM-Based Agents. \emph{arXiv:2507.07957}, 2025.

\bibitem{timem2026}
K. Li, X. Yu, Z. Ni, Y. Zeng, et al. TiMem: Temporal-Hierarchical Memory Consolidation for Long-Horizon Conversational Agents. \emph{arXiv:2601.02845}, 2026.

\bibitem{mem12025}
MEM1: Learning to Synergize Memory and Reasoning for Efficient Long-Horizon Agents. \emph{arXiv:2506.15841}, 2025.

\bibitem{amem2025}
W. Xu, K. Mei, Y. Zhang, et al. A-Mem: Agentic Memory for LLM Agents. \emph{arXiv:2502.12110}, 2025.

\bibitem{memoryagentbench2025}
MemoryAgentBench: Evaluating Memory in LLM Agents via Incremental Multi-Turn Interactions. \emph{arXiv:2507.05257}, 2025.

\bibitem{locomo2024}
A. Maharana, D.-H. Lee, S. Tulyakov, M. Bansal, F. Barbieri, and Y. Fang. Evaluating Very Long-Term Conversational Memory of LLM Agents. \emph{arXiv:2402.17753}, 2024.

\bibitem{longmemeval2024}
D. Wu, H. Wang, W. Yu, Y. Zhang, K.-W. Chang, and D. Yu. LongMemEval: Benchmarking Chat Assistants on Long-Term Interactive Memory. \emph{arXiv:2410.10813}, 2024.

\bibitem{amabench2026}
AMA-Bench: Evaluating Long-Horizon Memory for Agentic Applications. \emph{arXiv:2602.22769}, 2026.

\bibitem{memorysurvey2026}
Memory for Autonomous LLM Agents: Mechanisms, Evaluation, and Emerging Frontiers. \emph{arXiv:2603.07670}, 2026.

\bibitem{memagents2026}
MemAgents: Memory for LLM-Based Agentic Systems. ICLR 2026 Workshop Proposal. \emph{OpenReview id U51WxL382H}, 2026.

\bibitem{msftsemkernel}
Microsoft Semantic Kernel GitHub Issue \#13435: Deterministic execution and audit for agentic workflows. \emph{https://github.com/microsoft/semantic-kernel/issues/13435}, 2025.

\bibitem{sakuraskyaudit}
SakuraSky Trustworthy-AI Series: Audit-Ready Agents in Regulated Industries. \emph{sakurasky.com/trustworthy-ai}, 2025.

\bibitem{apistronghold}
API Stronghold. Audit and Compliance for LLM Agent Deployments. Industry whitepaper, 2025.

\bibitem{oracleagentmem}
Oracle AI Research Blog. Stateless Memory Substrates for Enterprise Agent Systems. \emph{oracle.com/ai/blog/agent-memory}, 2025.

\bibitem{realm2020}
K. Guu, K. Lee, Z. Tung, P. Pasupat, and M.-W. Chang. REALM: Retrieval-Augmented Language Model Pre-Training. \emph{Proceedings of the 37th International Conference on Machine Learning (ICML)}, 2020.

\bibitem{rag2020}
P. Lewis, E. Perez, A. Piktus, F. Petroni, V. Karpukhin, N. Goyal, H. Küttler, M. Lewis, W. Yih, T. Rockt\"aschel, S. Riedel, and D. Kiela. Retrieval-Augmented Generation for Knowledge-Intensive NLP Tasks. \emph{Advances in Neural Information Processing Systems (NeurIPS)}, 2020.

\bibitem{eventsourcing2005}
M. Fowler. Event Sourcing. \emph{martinfowler.com/eaaDev/EventSourcing.html}, 2005.

\bibitem{kleppmann2017}
M. Kleppmann. \emph{Designing Data-Intensive Applications}. O'Reilly Media, 2017. Chapter 11: Stream Processing.

\end{thebibliography}
\end{document}